\documentclass[10pt,twocolumn,letterpaper]{article}

\usepackage{cvpr}
\usepackage{times}
\usepackage{epsfig}
\usepackage{graphicx}
\usepackage{amsmath}
\usepackage{amssymb}

\usepackage{array}

\usepackage[pagebackref=false,breaklinks=true,letterpaper=true,colorlinks,bookmarks=false]{hyperref}

\cvprfinalcopy 

\def\cvprPaperID{1071} 

\ifcvprfinal\fi

\begin{document}

\title{Pooled Motion Features for First-Person Videos}

\author{M. S. Ryoo, Brandon Rothrock, and Larry Matthies\\
Jet Propulsion Laboratory, California Institute of Technology, Pasadena, CA\\
{\tt\small mryoo@jpl.nasa.gov}
}

\maketitle
\thispagestyle{empty}

\begin{abstract}
In this paper, we present a new feature representation for first-person videos. In first-person video understanding (e.g., activity recognition), it is very important to capture both entire scene dynamics (i.e., egomotion) and salient local motion observed in videos. We describe a representation framework based on time series pooling, which is designed to abstract short-term/long-term changes in feature descriptor elements. The idea is to keep track of how descriptor values are changing over time and summarize them to represent motion in the activity video. The framework is general, handling any types of per-frame feature descriptors including conventional motion descriptors like histogram of optical flows (HOF) as well as appearance descriptors from more recent convolutional neural networks (CNN).

We experimentally confirm that our approach clearly outperforms previous feature representations including bag-of-visual-words and improved Fisher vector (IFV) when using identical underlying feature descriptors. We also confirm that our feature representation has superior performance to existing state-of-the-art features like local spatio-temporal features and Improved Trajectory Features (originally developed for 3rd-person videos) when handling first-person videos. Multiple first-person activity datasets were tested under various settings to confirm these findings.

\end{abstract}


\section{Introduction}

\begin{figure}
	\centering
	\resizebox{1.0\linewidth}{!}{%
	  \includegraphics{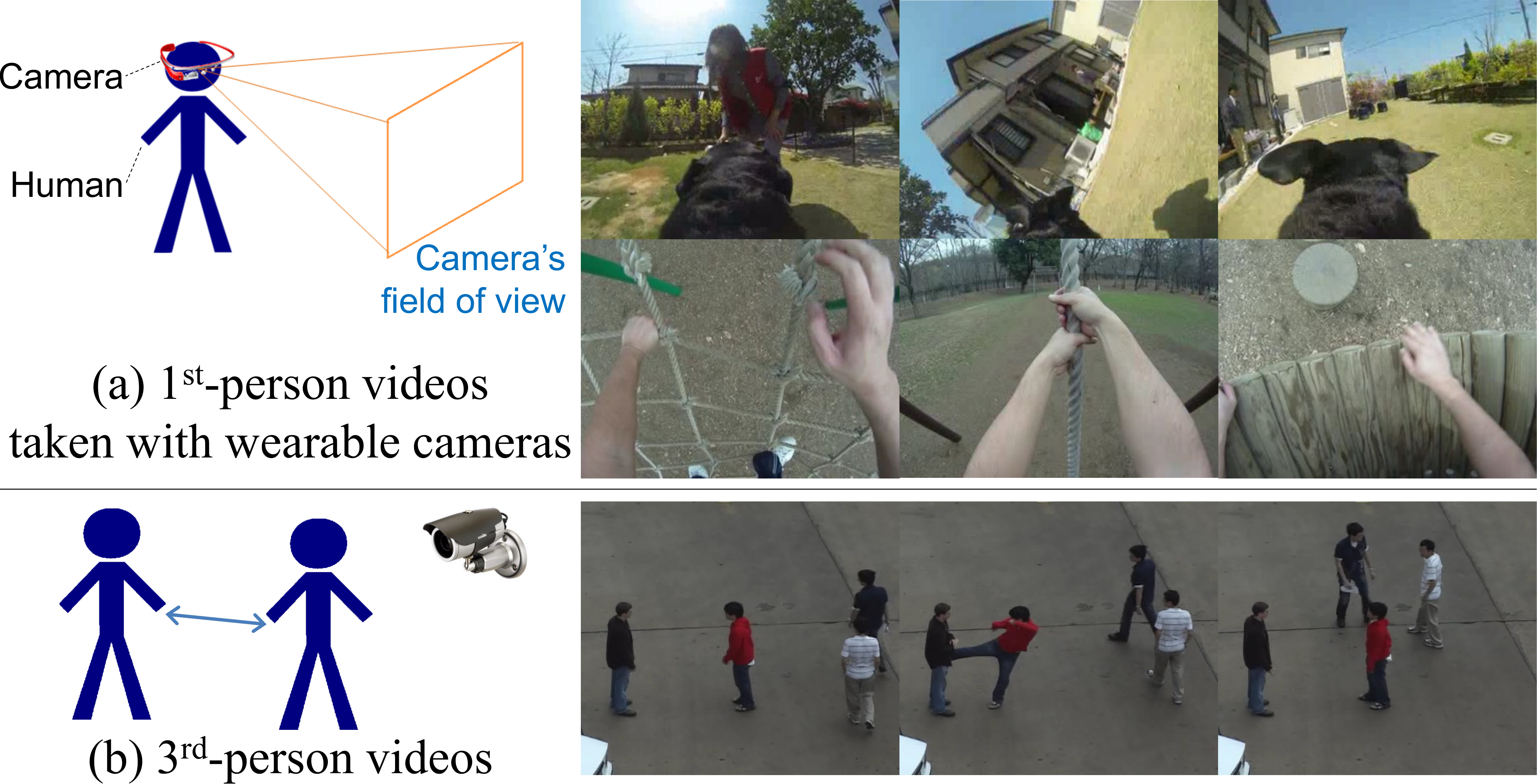}
	}
	\caption{Conceptual comparison between 1st-person videos and 3rd-person videos. Example snapshots from public first-person datasets \cite{kitani11,ryoo14dog} taken with human/animal wearable cameras and those from public 3rd-person dataset \cite{ryoo09iccv} are also illustrated.}
	\label{fig:intro}
\end{figure}

First-person videos, also called egocentric videos, are videos taken from an actor's own viewpoint. The volume of egocentric video is rapidly increasing due to the recent ubiquity of small wearable devices. The main difference between conventional 3rd-person videos and 1st-person videos is that, in 1st-person videos, the person wearing the camera is actively involved in the events being recorded. Strong egomotion is observed in first-person videos, which makes them visually very unique (Figure \ref{fig:intro}). Automated understanding of such videos (i.e., first-person activity recognition) is crucial for many societal applications including quality-of-life systems to support daily living and video-based life-logging. Applications also include robot perception and human-robot interactions, since videos from the robot's viewpoint naturally are in first-person.

Despite a massive amount of first-person videos becoming available, approaches to semantically understand such videos have been very limited. This is particularly true for research on `motion features' for first-person videos, which serves as a fundamental component for visual grounding of actions and events. Even though there has been previous works on extraction of first-person-specific semantic features like hand locations \cite{lee14} and human gaze \cite{li13}, features and representations designed to capture motion dynamics of first-person videos have been lacking. Representing this egomotion is very essential for recognition of sports activities, accident activities for patient/health monitoring (e.g., a person collapsing), activities for surveillance/military (e.g., another person assaulting), and many others from first-person videos. Most of the previous first-person activity recognition works \cite{kitani11,ryoo13} focused on the use of existing features and representations designed for conventional 3rd-person videos, without tailoring motion features for the first-person case.



This paper introduces a new feature representation named \emph{pooled time series} (PoT). Our PoT is a general representation framework based on time series pooling of feature descriptors, which is particularly designed to capture motion information in first-person videos. Given a sequence of per-frame feature descriptors (e.g., HOF or CNN features) from a video, PoT abstracts them by computing short-term/long-term changes in each descriptor element. The motivation is to develop a new feature representation that captures `details' of entire scene dynamics displayed in first-person videos, thereby obtaining better video recognition performances. Capturing egomotion information is crucial for recognition of ego-actions and interactions from first-person videos, and our PoT representation allows the system to do so by keeping track of very detailed changes in feature descriptor values while suppressing noise. Multiple novel pooling operators are introduced, and are combined with temporal filters to handle the temporal structure of human activities.

We experimentally confirm that our proposed PoT representation clearly outperforms previous feature representations such as bag-of-visual-words and improved Fisher vector \cite{ifv10} on first-person activity recognition. Both the global motion aspect and local motion aspect of first-person videos are captured with our PoT by taking advantage of different types of descriptors, and we illustrate recognition accuracies of our PoT with each of the descriptors as well as their combinations. Furthermore, we demonstrate that our combined PoT representation has superior performance to the best-known motion feature designed for 3rd-person videos \cite{wang13}, when handing 1st-person videos.

\subsection{Related works}

Recognition from first-person videos is a topic with an increasing amount of attention. There are works focusing on first-person-specific features, including hand locations in first-person videos \cite{lee14} and human gaze estimation based on first-person videos \cite{li13}. There also have been works on object recognition from first-person videos \cite{lee12,ramanan12}.

However, study on motion features for first-person videos has been relatively limited, particularly those for first-person activity recognition. Most of the works focused on temporal segmentation of videos using optical flow-based features, without taking advantage of high-dimensional image features for detailed recognition of high-level activities. Kitani \emph{et al.} \cite{kitani11} worked on unsupervised learning of ego-actions and segmentation of videos based on it. A simple histogram based on optical flow direction/magnitude and frequency was constructed as a feature representation, which can be viewed as an extension of HOF. Poleg \emph{et al.} \cite{poleg14} introduced the use of displacement vectors similar to optical flows for long-term temporal segmentation, but they only focused on segmentation of relatively simple egomotion such as walking and wheeling. \cite{ryoo13} investigated the first-person activity recognition scenarios by combining multiple features, while particularly focusing on recognition of interaction-level activities. Still, they used very conventional HOF and local spatio-temporal features \cite{laptev05,dollar05} together with general bag-of-visual-words representation, without any attempt to develop first-person-specific features.





\section{Pooled times series representation}
\label{sec:pot}

In this section, we introduce our new feature representation named \emph{pooled time series} (PoT), which is specifically designed for first-person videos. The role of a feature `representation' is to abstract a set of raw feature descriptors (e.g., histogram of oriented gradients) extracted from each video into a single vector representing the video. It converts a large number of high-dimensional descriptors into a single vector with a tractable dimensionality, allowing its result to serves as an input vector for classifiers (e.g., activity classification). Existing feature representations include bag-of-visual-words (BoW) and improved Fisher vector (IFV), which converts a set of raw descriptors into a low-dimensional histogram. What we introduce in this section is a new feature representation that better abstracts motion displayed in first-person videos.

\begin{figure*}
	\centering
	\resizebox{1.0\linewidth}{!}{%
	  \includegraphics{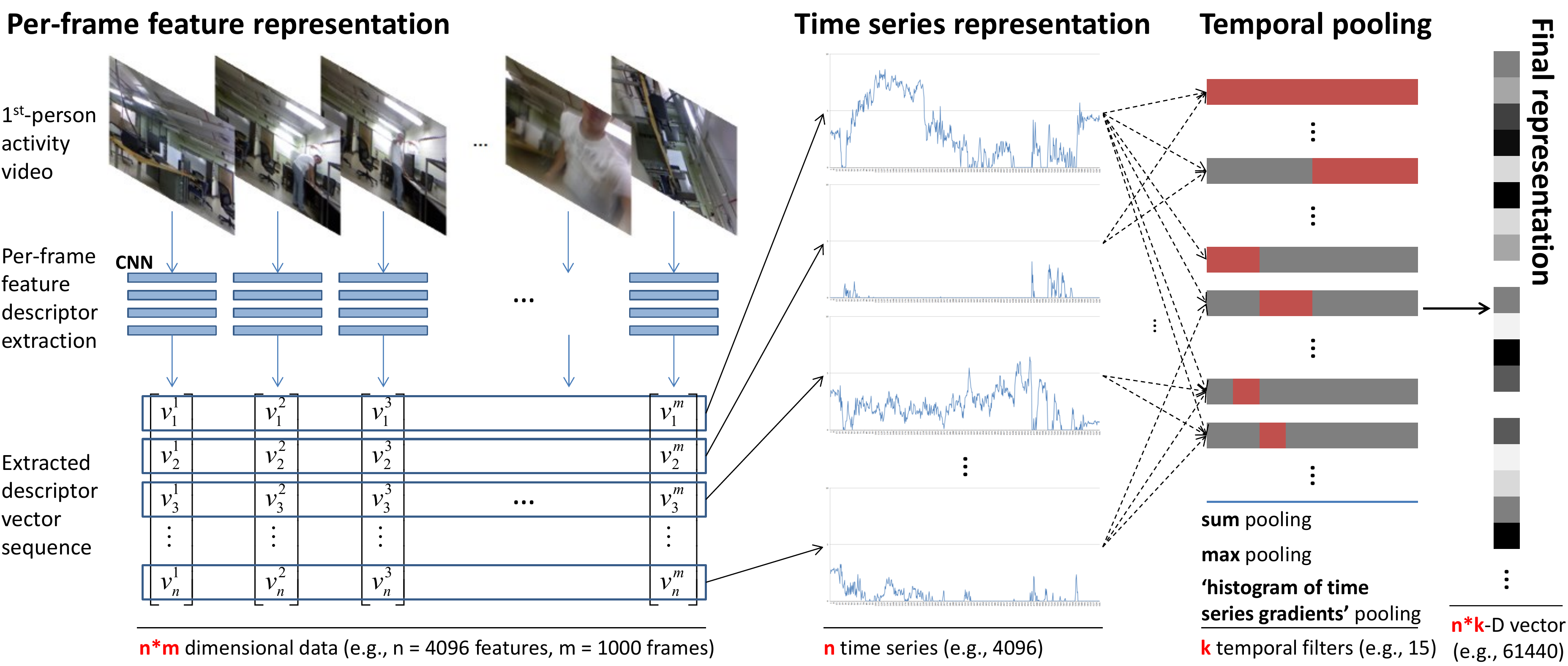}
	}
	\caption{Overall representation framework of our pooled time series (PoT).}
	\label{fig:representation}
\end{figure*}

The overall pipeline of our PoT representation is as follows. Given a first-person video (i.e., a sequence of image frames), our approach first extracts appearance/motion descriptors from each frame. As a result, a sequence of $n$-dimensional descriptor vectors is obtained where $n$ is the size of the vector from each frame. Our approach interprets this as a set of $n$ time series. The idea is to keep track of how each element of the descriptor vector is changing over time (i.e., it becomes a function of time), and summarize such information to represent the activity video. Next, temporal pooling is performed: a set of temporal filters (i.e., time intervals) is applied to each time series and the system performs multiple types of pooling operations (e.g., max, sum, gradients, ...) per filter. Finally, the pooling results are concatenated to form the final representation of the video. Figure \ref{fig:representation} illustrates the overall process.

Let each per-frame feature descriptor obtained at frame $t$ be denoted as $V^t = [v_1^t, v_2^t, ..., v_n^t]$. Our PoT representation framework interprets this sequence of vectors $V^1, ...,V^m$ ($m$ is the number of video frames) as a set of time series, $\{f_1(t), ..., f_n(t)\}$. That is, each of our time series $f_i(t)$ corresponding to the $i$th feature descriptor value is defined as $f_i(t) = v_i^t$. For each time series, temporal pooling is performed with a set of $k$ temporal filters, which essentially is a set of time intervals to make the system focus on each local time window: $\{[t_1^s, t_1^e], ..., [t_k^s, t_k^e]\}$.  A temporal pyramid structure \cite{choi08} is used in our implementation to obtain filters, but any number of filters with (overlapping) intervals can be used by our framework in principle.

Finally, multiple pooling operators are applied for each filter and their results are concatenated to obtain the final PoT feature representation of the video:
\begin{equation}
	\begin{aligned}
		x = \left[ x_1^{op_1}[t_1^s, t_1^e], ~x_1^{op_2}[t_1^s, t_1^e], \cdots,  ~x_n^{op_r}[t_k^s, t_k^e]  \right]
	\end{aligned}
\end{equation}
where $x_i^{{op}_j}$ specifies that it is applying the $j$th pooling operator to the $i$th time series $f_i(t)$. Our PoT representation takes advantage of four different types of pooling operators including two newly introduced temporal pooling operators, which we discuss more in Subsection \ref{subsec:pooling}.

Our framework of (i) extracting per-frame descriptors, (ii) interpreting them as a set of time series, and (iii) applying various types of time series pooling and concatenating them provides the following three important abilities:

 
First, (1) it preserves detailed dynamics displayed in each descriptor element as a time series, and allows the representation to capture both long-term motion and short-term information with multiple temporal filters. That is, depending on the nature of the time series, our representation is able to capture subtle short-term motion by pooling from a filter with a small time interval as well as long-term motion by performing pooling with a large time interval. Such flexibility is in contrast to previous bag-of-visual-words representation for global motion descriptors (e.g., the one used in \cite{ryoo13}) that abstracts all descriptor values in one frame (or a subsequence of few frames) into a single discretized `visual word'. In addition, (2) our representation explicitly imposes temporal structure of the activity by decomposing the entire time interval to multiple subintervals, which is very important for representing high-level activities. Finally, (3) it allows the system to take advantage of multiple types of pooling operators so that the representation captures different aspects of the data.


As a result of our framework, each video is represented with one single vector having a tractable dimensionality. Activity recognition is performed by training/testing standard classifiers (e.g., SVM classifiers) with these vectors. Our representation is able to cope with any type of generative and discriminative classifiers in principle, and we show its superiority over others in Section \ref{subsec:representation-eval}.


\subsection{Handling high-dimensional feature descriptors}

The proposed representation framework is very general in the aspect that it is able to cope with any types of per-frame image/motion descriptors such as histogram of oriented gradients (HOG) or histogram of optical flows (HOF). Furthermore, it is particularly designed to handle high-dimensional per-frame image descriptors: image-based deep learning features which are also called convolutional neural network (CNN) features \cite{jia14caffe,overfeat14}. These deep learning image features are obtained by snatching intermediate outputs from internal convolutional layers of a CNN, pre-trained on image datasets. They can also be viewed as cascades of automatically learned image filters. These image descriptors are trained from large scale image datasets and have obtained highly successful results on image classification/detection \cite{girshick13} as well as video classification \cite{karpathy14}, performing superior to state-of-the-art hand-designed image descriptors such as HOG even without re-training the networks.


Our motivation was to design a general representation that best takes advantage of such high-dimensional descriptors and confirm that these CNN features are able to increase first-person activity recognition performance significantly together with other features. Each element of a CNN feature vector abstracts particular local/global appearance for a single frame, and (by extension) its time series models how this local/global appearance is changing over time. As a human in the scene moves (e.g., changes his/her posture) and the camera changes its viewpoint because of egomotion, certain CNN feature values will become activated/deactivated and our idea is to keep track of such changes to represent the activity video. In our experiments, we explicitly confirm this while comparing our representation with the conventional representations. When using CNN features as our base per-frame descriptors, we get a 4096-dimensional feature vector (i.e., n=4096) for each image frame by obtaining outputs of the last convolutional layer (e.g., stage 7 in \cite{overfeat14}). 


\subsection{Temporal pooling operators}
\label{subsec:pooling}

Our PoT representation is constructed by applying multiple types of temporal pooling operators over each temporal filter (i.e., time interval). In this paper, we take advantage of four different types of pooling operators: conventional max pooling and sum pooling, and two types of our new `histogram of time series gradients' pooling.

Our max pooling and sum pooling operators are defined as follows:
\begin{equation}
	\begin{aligned}
		x_i^{\max}[t^s, t^e] =& \max_{t = t_s..t_e} f_i(t),\\
		x_i^{\sum}[t^s, t^e] =& \sum_{t = t_s}^{t_e} f_i(t).
	\end{aligned}
\end{equation}

In addition to these traditional pooling operators, we newly introduce the concept of `histogram of time series gradients' pooling. The idea is to count the number of positive (and negative) gradients within the temporal filter.
\begin{equation}
	\begin{aligned}
		x_i^{\Delta_1^+} [t^s, t^e] &= |\{t ~|~ f_i(t) - f_i(t-1) > 0 \wedge t^s \leq t \leq t^e\}|,\\
		x_i^{\Delta_1^-} [t^s, t^e] &= |\{t ~|~ f_i(t) - f_i(t-1) < 0 \wedge t^s \leq t \leq t^e\}|.
	\end{aligned}
\end{equation}
Furthermore, we propose a variation of the above new pooling operator, which sums the amount of positive (or negative) gradients instead of simply counting their numbers. It is defined as:
\begin{equation}
	\begin{aligned}
		x_i^{\Delta_2^+} [t_s, t_e] = \sum_{t = t_s}^{t_e} h^+_i(t), ~~x_i^{\Delta_2^-} [t_s, t_e] = \sum_{t = t_s}^{t_e} h^-_i(t)
	\end{aligned}
\end{equation}
where 
\begin{equation}
	\begin{aligned} 
		 h^+_i(t) &= \left\{ \begin{array}{cl}
					f_i(t) - f_i(t-1) &\mbox{ if $(f_i(t) - f_i(t-1))>0$} \\
					0 &\mbox{ otherwise},
     \end{array} \right.\\
     h^-_i(t) &= \left\{ \begin{array}{cl}
					f_i(t-1) - f_i(t) &\mbox{ if $(f_i(t) - f_i(t-1))<0$} \\
					0 &\mbox{ otherwise}.
     \end{array} \right.
	\end{aligned}
\end{equation}
Each of our time series gradients pooling operation generates a pair of values (i.e., $x_i^{\Delta^+}$ and $x_i^{\Delta^-}$) instead of a single value like max pooling. These two values are concatenated in our PoT representations.

\section{Experiments}

\subsection{Experimental settings}

{\flushleft\textbf{Datasets:} We conducted our experiments with two different public first-person video datasets: DogCentric activity dataset \cite{ryoo14dog} and UEC Park dataset \cite{kitani11}. These are very challenging datasets with strong camera egomotion, which are different from conventional 3rd-person datasets. Figure \ref{fig:intro} shows sample images. DogCentric dataset was recorded with wearable cameras mounted on dogs' back. UEC Park dataset was collected by a human wearing a camera. DogCentric dataset contains ego-actions of the dog as well as interactions between the dog and other humans (e.g., a human throws a ball and the dog chases it). UEC Park dataset contains ego-actions of the person (wearing a camera) involved in various types of physical activities (e.g., climbing a ladder) at a park. Dogcentric dataset consists of 10 activity classes, while UEC dataset consists of 29 classes.}

\vspace{-2pt}
{\flushleft\textbf{Representation implementation:} We implemented our PoT representations with four different types of per-frame feature descriptors: histogram of optical flows (HOF), motion boundary histogram (MBH) used in \cite{wang13}, Overfeat CNN image feature \cite{overfeat14}, and Caffe CNN image feature \cite{jia14caffe}. The first two descriptors (i.e., HOF and MBH) are optical flow based motion descriptors, and the last two (i.e., Overfeat and Caffe) are deep learning based image appearance descriptors from CNNs pre-trained on ImageNet. Our HOF descriptors are in 200-D (5-by-5-by-8), MBH descriptors are in 400-D (two 5-by-5-by-8), and Overfeat and Caffe are in 4096-D. L1 normalization was applied for each descriptor. As a result, four different versions of our PoT representations were implemented as well as the final representation combining all four representations. As described in Section \ref{sec:pot}, pyramid temporal filters with level 4 were used and four different types of pooling operators were applied.}


\vspace{-2pt}
{\flushleft\textbf{Classifiers:} In all our experiments, we used the same non-linear SVM with a $\chi^2$ kernel. It showed better performance compared to linear SVM. When combining representations with multiple descriptors, a multi-channel kernel was used.}

\vspace{-2pt}
{\flushleft\textbf{Evaluation setting:} We followed the standard evaluation setting of the DogCentric dataset: we performed repeated random training/testing splits 100 times, and averaged the performance. We randomly selected half of videos per activity class as training videos, and used the others for the testing. If the number of total videos per class is odd, we made the testing set to contain one more video. Once training videos are selected, they are used across the entire experiments for fair comparisons.}

\subsection{Feature representation evaluation}
\label{subsec:representation-eval}

We conducted experiments to confirm superiority of our proposed PoT representation over conventional feature representations. The idea is to evaluate the performances of our PoT, and compare it with those of other widely-used state-of-the-art representations while making them use exactly the same feature descriptors. More specifically, we compared our PoT representations with bag-of-visual-words (BoW) and Improved Fisher Vector (IFV) \cite{ifv10} while using four types of feature descriptors (i.e., HOF, MBH, Overfeat, and Caffe) and their combinations.

BoW and IFV are very commonly used feature representations in the activity recognition literature \cite{aggarwal11}. BoW represents a video as a histogram of `visual words' by clustering all feature descriptors, making each descriptor assigned to a specific visual word. IFV can be viewed as a `soft' version of that, in the aspect that it represents each descriptor as a set of soft assignments to cluster centers. We tested tens of different parameter settings for each feature type, and chose the best setting per feature. This includes the tuning of the number of visual words (e.g., IFV has 4000-D for HOF and 40960-D for Caffe). In addition, we implemented BoW and IFV in conjunction with the temporal pyramid pooling identical to the one used in our PoT, so that they also consider temporal structure among features. We explicitly compared all these different representations with our PoT. Also, since the clustering processes of BoW and IFV contain randomness, we report their 95\% confidence interval together with the median performance by testing them 10 times.


\begin{figure}
	\centering
	\resizebox{1.0\linewidth}{!}{
	  \includegraphics{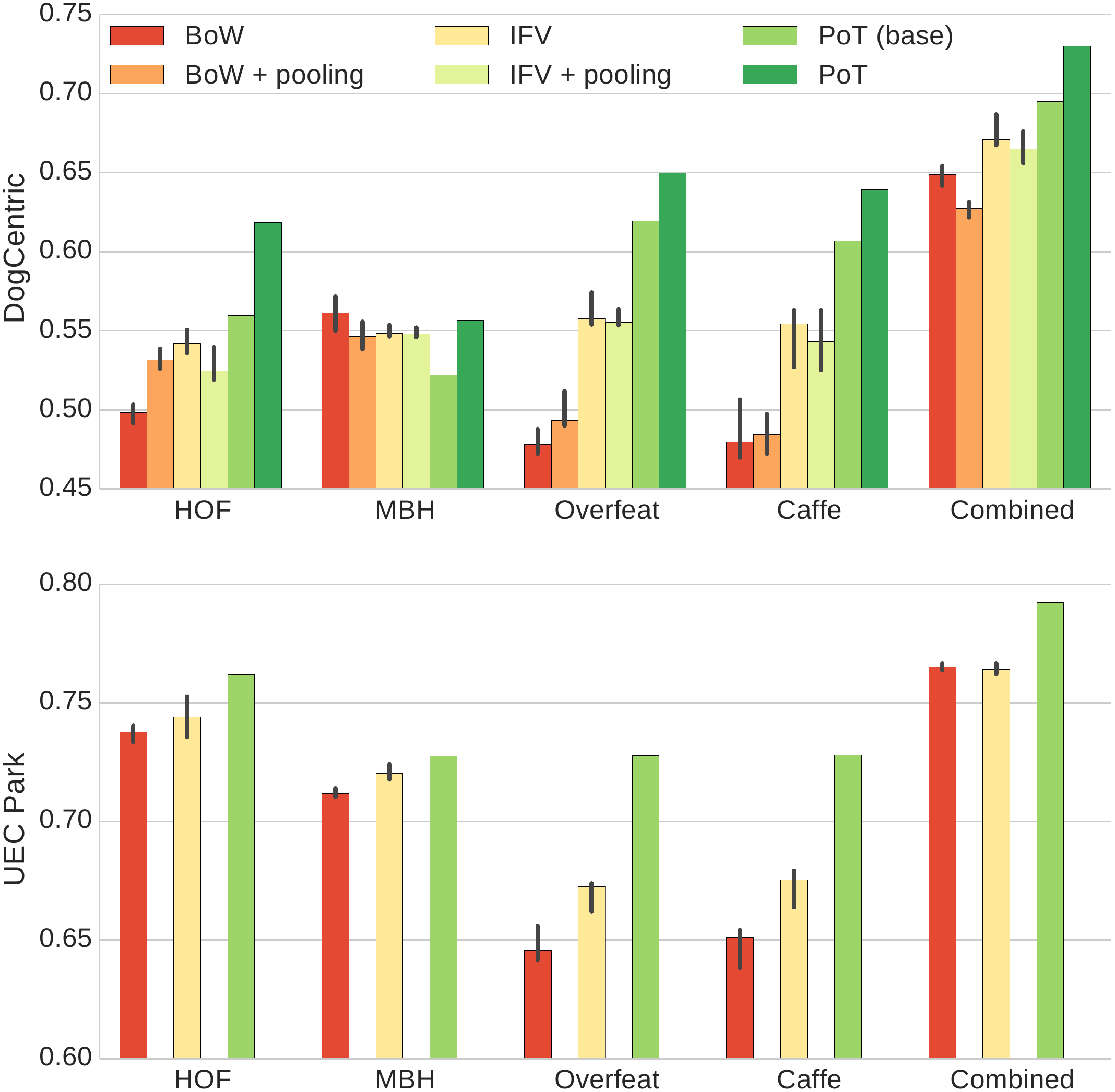}
	}
	\caption{Classification accuracies of feature representations with each descriptor (and their combination). Representations that utilize randomness are drawn with 95\% confidence intervals. See text for details.}
	\label{fig:performance}
\end{figure}

\vspace{-2pt}
{\flushleft\textbf{DogCentric activity dataset:} Figure \ref{fig:performance} (top) describes the 10-class activity classification accuracies of our representation (and BoW and IFV) for each of the base descriptors.}

Here, we are showing the accuracies of the PoT representation with the best combination of pooling operators. As described in Section \ref{subsec:pooling}, there are four different pooling operators our PoT representation can take advantage of. We conducted experiments with all possible combinations of pooling operators for PoT (which can be found in our supplementary Appendix), and selected the best performing combination. In general, concatenations of all pooling operators (e.g., $\sum$+$\max$+$\Delta_1$) obtained the best results, or results very close to the best. `PoT (base)' is the basic version of our feature representation, which is constructed by applying the pooling operator to a single time interval that covers the entire activity video (i.e, no temporal pyramid structure).


We are able to observe that our PoT representations perform superior to BoW and IFV in all cases, except for MBH descriptors where all representations showed similar performances. Even when we add temporal pyramid pooling (identical to the one used in our representation) to BoW and IFV, their performances were clearly inferior to our PoT. The mean accuracy of the combined IFV representation (with pyramid) was 0.666, while our PoT obtained the accuracy of 0.730. Previous state-of-the-art is 0.605 \cite{ryoo14dog}.

Particularly, in the case of using high-dimensional deep learning features (i.e., 4096-D in Overfeat and Caffe), we confirmed that our representation significantly improves the performance over both BoW and IFV. We believe this is due to the fact that per-frame abstraction (i.e., clustering) performed in BoW and IFV fails to capture subtle local cues while our representation is particularly designed to handle such high-dimensionality descriptors. BoW abstracts each high-dimensional per-frame descriptor into a single `visual word' (or a few soft assignments in case of IFV). As a result, subtle changes in a small number of descriptor values are ignored in general, which is particularly harmful in the case of high-dimensional descriptors. This is unlike our PoT that tries to explicitly capture such changes with time series pooling. The result suggests that PoT is the better representation to take advantage of modern high-dimensional feature descriptors.

Another important observation is that our PoT benefitted greatly by considering temporal structures among features, much more compared to BoW and IFV. We discuss this more in Subsection \ref{subsec:temporal}.

In addition, since dynamic time warping (DTW) is a traditional approach to deal with time series data, we also tested a basic DTW-based template matching (using the same time series with PoT) as a baseline. The best performance of DTW was 0.288, as opposed to 0.730 of ours.

\vspace{-2pt}
{\flushleft\textbf{UEC Park dataset:} We also performed the same experiments described above with one more public first-person video dataset: UEC Park dataset. As described in the previous subsection, the dataset contains video segments labeled with 29 different classes. Labels are very rough and the number of videos per activity class are very unbalanced in this dataset (e.g., there is a class with only 1 video), making the classification task challenging.}

Also, videos of this dataset were obtained by segmenting a long video every 2 seconds, and each segment was labeled based on the most dominant activity observed in the segment. As a result, activities in these videos are not temporally aligned (e.g., a video segment may not even contain the initial part of the activity at all) and using pooling with temporal structures only harms the recognition performances. We confirmed this with all representations: BoW, IFV, and PoT. Thus, here, we only show the results of representations without any temporal structure consideration. PoT is at a disadvantage for this dataset, since it benefits greatly using pooling with temporal structures while BoW and IFV do not, as we discuss more in Subsection \ref{subsec:temporal}.

Figure \ref{fig:performance} (bottom) shows the result. Our PoT obtained the best performance on all feature descriptors, similar to the case of the DogCentric dataset. PoT obtained the best result using all four feature descriptors and obtained particularly higher performances for high-dimensional CNN features. Even though PoT was not able to fully take advantage of activites' temporal structures, it still performed superior to BoW and IFV.

\begin{table*}
	\caption{A table comparing performances of the proposed approach with state-of-the-arts on DogCentric dataset \cite{ryoo14dog}: F1-scores per class and the final 10-class classification accuracies. Approaches with our representations are colored blue. The performances are split into three categories: representations with only one descriptor, representations with multiple descriptors, and combinations of multiple different features representations. The best performance per category is indicated with bold. The overall best performance is indicated with bold+red.}
	\label{table:comparison-dog}

	\centering
	\resizebox{1.0\linewidth}{!}{%
	  \includegraphics{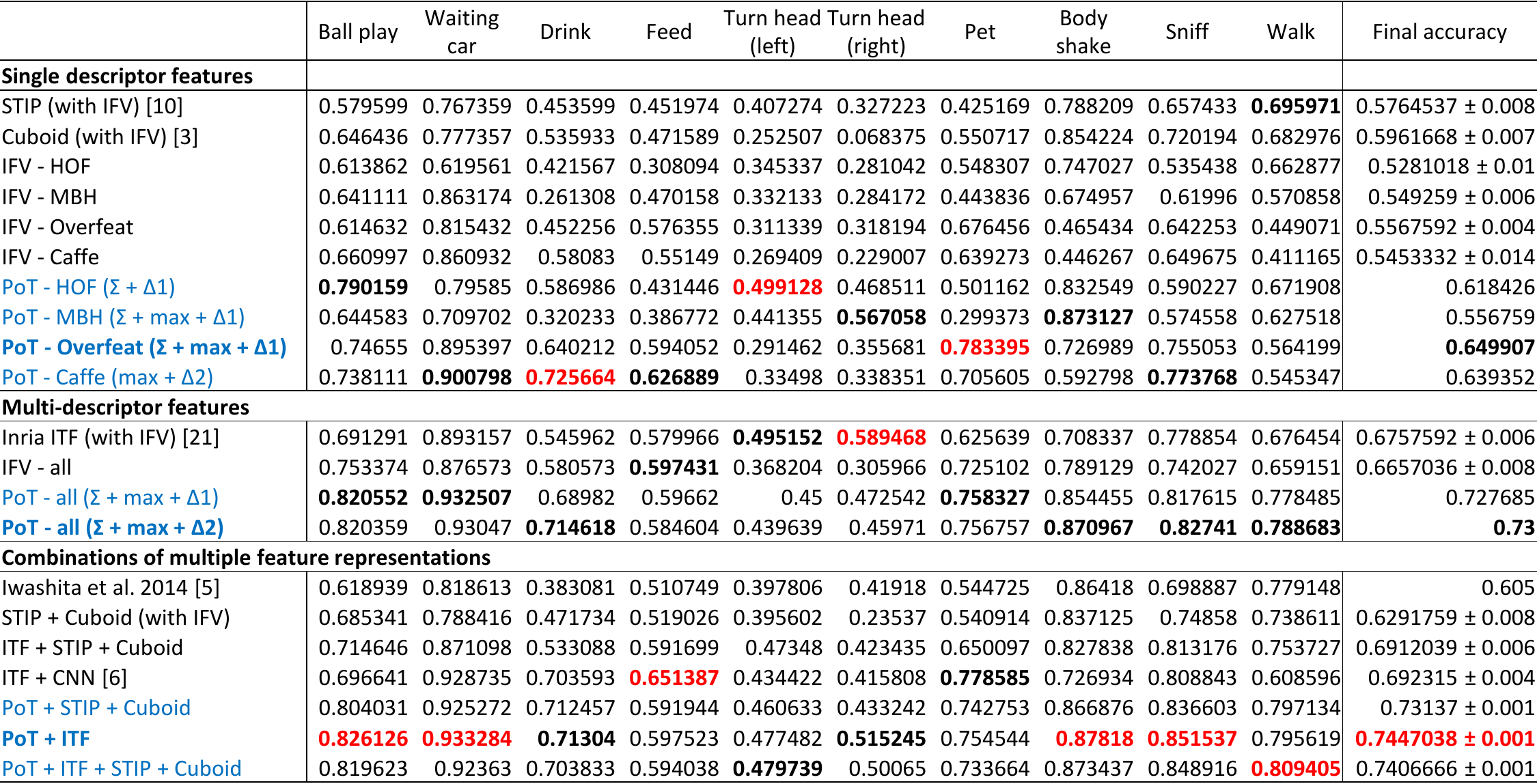}
	}
\end{table*}

\subsection{Temporal structure evaluation}
\label{subsec:temporal}

We conducted further experiments to confirm the advantage of PoT: it benefits more compared to other representations when considering temporal structure among features. DogCentric activity dataset was used for this experiment. We illustrate classification accuracies of BoW, IFV, and PoT with and without consideration of the temporal pyramid structure. `Without pyramid' means that the feature representation is constructed by applying the pooling operator on one single time interval that covers the entire activity video. `With pyramid' means that a set of temporal filters were used. For PoT, we also compare results of different time series pooling operators (and their combinations) with and without temporal pyramids.


\begin{figure}
	\centering
	\resizebox{1.0\linewidth}{!}{
	  \includegraphics{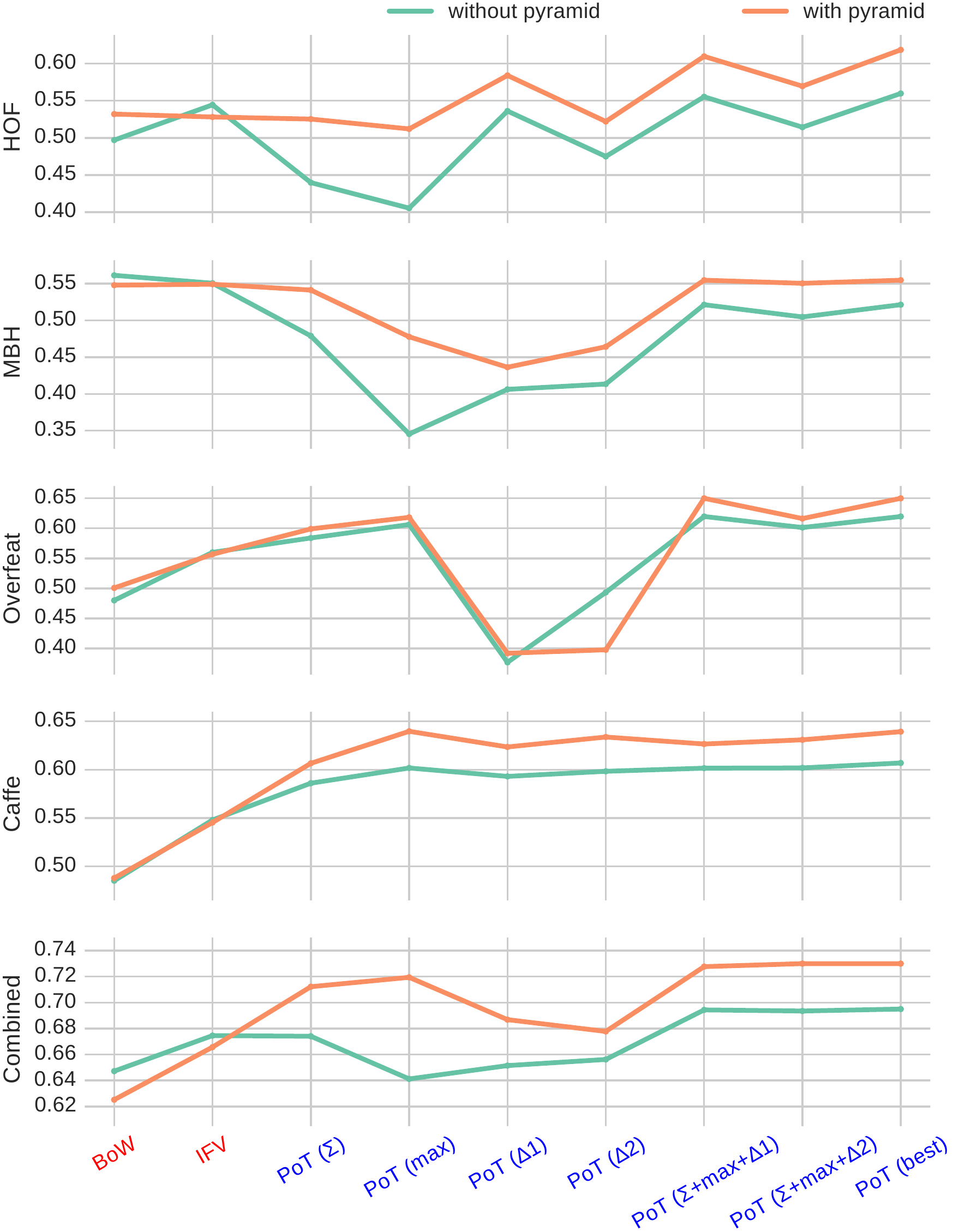}
	}
	\caption{Feature performance using BoW and IFV compared with various PoT pooling operators with and without a temporal pyramid on the DogCentric dataset. Y-axis is classification accuracy, and X-axis shows different representations. PoT generally benefits much more from the temporal pyramid than BoW and IFV.}
	\label{fig:dogcentric-pooling-chart}
\end{figure}

Figure \ref{fig:dogcentric-pooling-chart} shows the results. We are able to confirm that consideration of temporal structure benefited the recognition with our PoT while it did not benefit the other representations much. This is particularly true for the representations with combinations of all four descriptors. We believe this observation is caused by the following characteristic: as mentioned in Subsection \ref{subsec:representation-eval}, abstraction/discretization of per-frame observations in BoW (or IFV) completely ignores subtle local descriptor changes. This makes them have less chance to capture short-term local motion even when we temporally split the video using a pyramid. On the other hand, PoT does not suffer from such abstraction.



\subsection{Comparison to state-of-the-art features}

We explicitly compared activity classification accuracies of our PoT representations with other state-of-the-art features, including well-known local spatio-temporal features \cite{laptev05,dollar05} and recent trajectory-based local video features \cite{wang13}. Notably, INRIA's improved trajectory feature (ITF) \cite{wang13} is the one that obtained the best performance in the ICCV 2013 challenge on UCF101 dataset \cite{ucf101}. ITF internally takes advantage of three different feature descriptors similar to ours: HOG, HOF, and MBH. IFV representations were used for all these features. For the DogCentric dataset experiments, the temporal pyramid structure was considered by all of these previous features, since using temporal pyramid improved their classification accuracies by 0.03$\sim$0.05. What we show is that our new feature representations (together with frame-based descriptors) are more suitable for representing motion in first-person videos compared to the previous features designed for 3rd-person videos.


In addition, we implemented the approach of combining ITF with CNN descriptors \cite{jain14}, which won the ECCV 2014 classification challenge on UCF101 dataset. Both Overfeat and Caffe were used, and mean of per-frame CNN vectors were computed and added to ITF.



\begin{figure}
	\centering
	\resizebox{1.0\linewidth}{!}{
	  \includegraphics{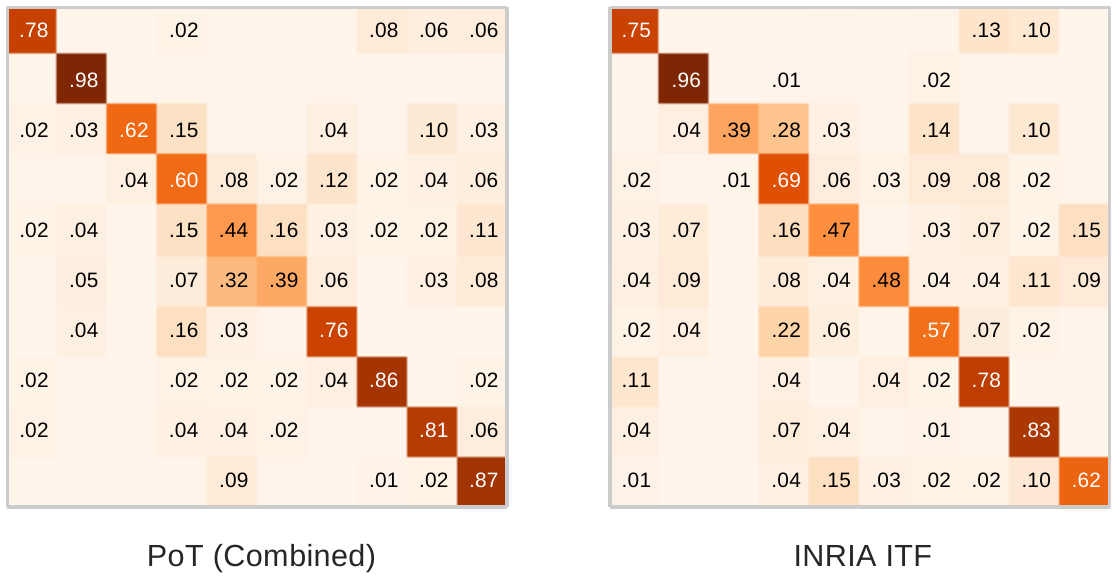}
	}
	\caption{Confusion matricies on the DogCentric dataset, comparing our PoT (combined) and the state-of-art INRIA ITF feature (with IFV).}
	\label{fig:confusion-dog}
\end{figure}

Table \ref{table:comparison-dog} shows the results with the DogCentric dataset. In addition to the final 10-class classification accuracies of the approaches, we are also reporting per-class F1-scores of them. The motivation is to analyze which feature descriptor/representation is better for recognition of which activity class. Instead of simply reporting per-class classification accuracies that do not take false positives into account, we computed a pair of precision and recall values per class from the confusion matrix and obtained F1-scores based on them.

\begin{table}
	\caption{A table comparing performances of the proposed approach with state-of-the-arts using UEC Park dataset \cite{kitani11}: 29-class classification accuracies are shown.}
	\label{table:comparison-uec}

	\centering
	\resizebox{0.75\linewidth}{!}{%
	  \includegraphics{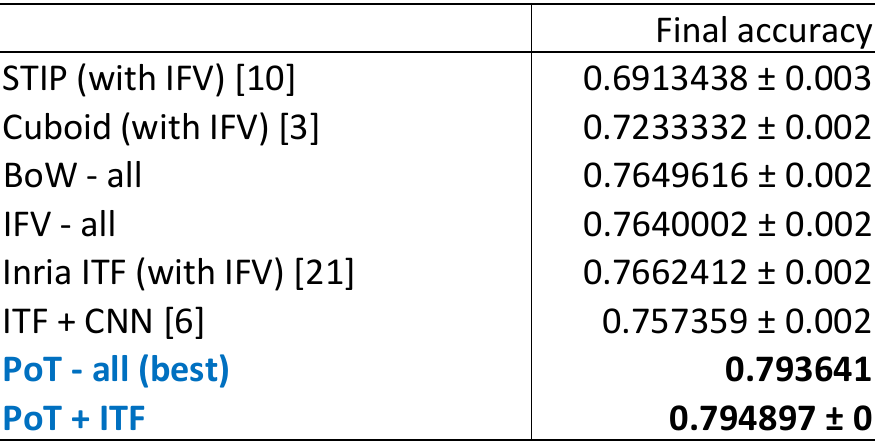}
	}
\end{table}

The result clearly illustrate that our PoT obtains the best result, outperforming local spatio-temporal features as well as the previously reported results \cite{ryoo14dog}. Particularly, our PoT performed significantly superior to the state-of-the-art ITF approach. Even with the temporal pyramid pooling added to the original ITF, our PoT performed much better than the ITF: 0.676 vs. 0.730. The ITF performance without pyramid was 0.638. Our PoT also showed the best per-class recognition accuracies in most of the classes. ITF showed slightly better performances for `turn head' classes, since these videos are actually more similar to 3rd-person videos: the camera was mounted on the back of the dog (i.e., not head) and these `turn head' videos do not involve much camera motion unlike the others. Furthermore, our PoT performed superior to the conventional method \cite{jain14} of combining ITF and mean per-frame CNN: 0.692 vs. 0.730.

We are able to observe similar results with the UEC Park dataset. Table \ref{table:comparison-uec} shows the results. PoT obtained the best performance, and we were able to (slightly) increase the performance further by combining PoT with ITF.

\subsection{Evaluation of appearance-based features}

\begin{figure}
	\centering
	\resizebox{1.0\linewidth}{!}{
	  \includegraphics{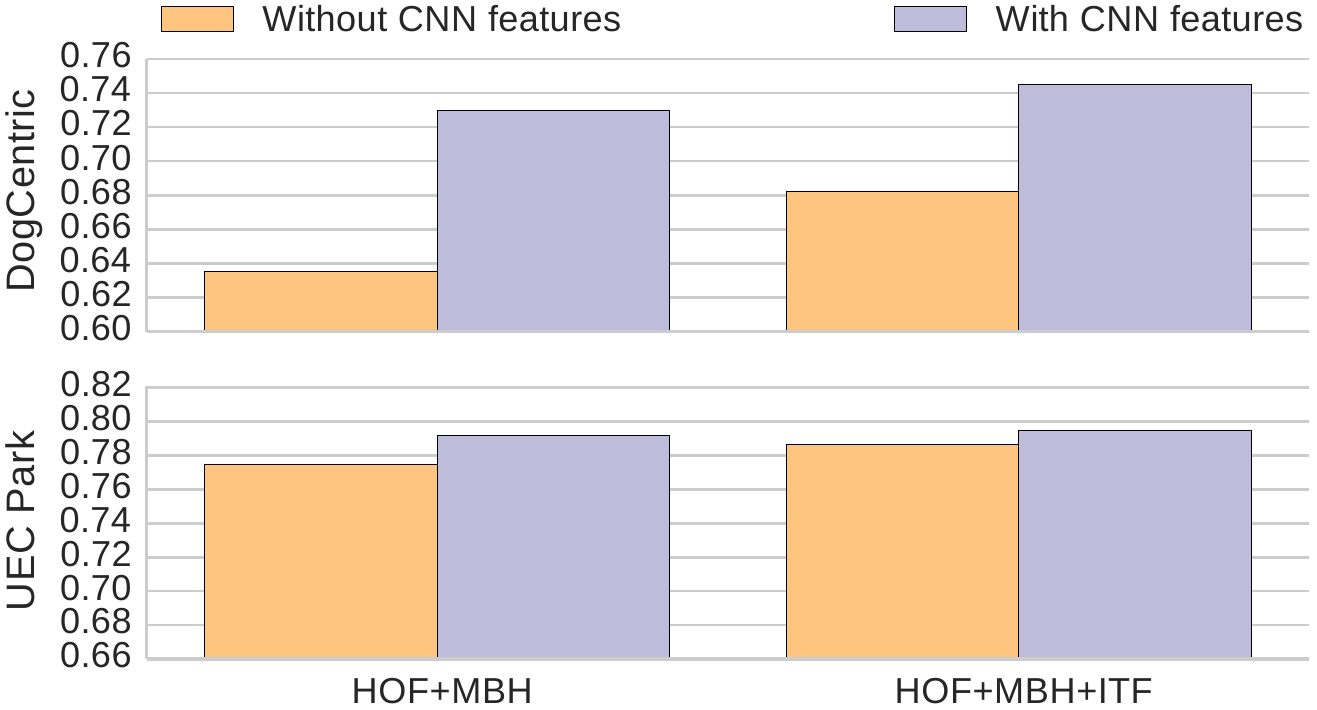}
	}
	\caption{Performance gain from combining CNN features with conventional motion features for both datasets.}
	\label{fig:cnn-contribution}
\end{figure}

{\flushleft\textbf{Taking advantage of CNN descriptors:} We explicitly compared the recognition accuracies of our approach with and without CNN-based appearance descriptors. The idea was to confirm `how much benefit our PoT representation is able to get from CNN descriptors' when representing first-person videos for activity recognition. The result is that, with our PoT, CNN-based appearance descriptors capture information different from motion descriptors and combining them with others really benefits the entire system, while the degree of their effectiveness is dependent on the dataset and activities. Figure \ref{fig:cnn-contribution} shows the results.}


For the DogCentric dataset, using CNN descriptors greatly benefited the overall recognition accuracy. Notice that CNN descriptors themselves showed superior performance compared to all other motion-based descriptors (e.g., HOF) with our PoT, as described in Figure \ref{fig:performance}. DogCentric dataset contains activity videos taken at various environments (indoor, outdoor, ...), and certain activities are highly correlated with such environment/background information (e.g., there will not be `ball chasing' activity in an indoor environment). As a consequence, capturing appearance information is very important for these activities/videos, and CNN descriptors showed very good results on them with our PoT. On the other hand, all UEC Park video sequences are taken at a same environment (i.e., a park), and thus CNN-based appearance descriptors were not as effective as motion descriptors. Nevertheless, in both cases, combining CNN features with other descriptors benefited the overall recognition performances, suggesting that our PoT is properly taking advantage of them.

\begin{figure}
	\centering
	\resizebox{1.0\linewidth}{!}{
	  \includegraphics{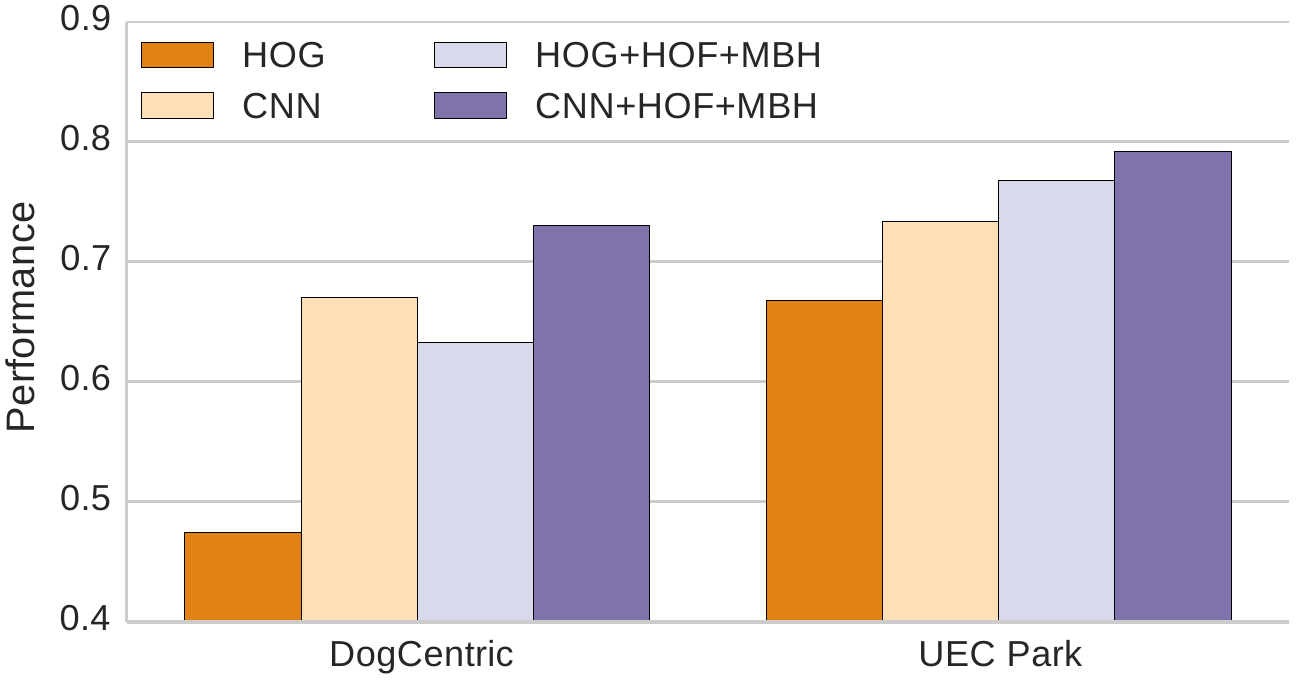}
	}
	\caption{Performance comparison of CNN features as a replacement for HOG, showing consistent gains on both datasets with and without motion features included.}
	\label{fig:cnn-vs-hog}
\end{figure}

{\flushleft\textbf{Appearance descriptors: CNN vs. HOG:} We tested another appearance descriptor, histogram of oriented gradients (HOG), and compared it with the CNN descriptors we are using. For this experiment, we extracted pure histogram of oriented gradients similar to our HOF from images. For each frame, a HOG descriptor was constructed with 8 different gradient directions and 5-by-5 spatial bins. Then, each sequence of these HOG descriptors was represented using our PoT representation. We not only compared our PoT representation only based on HOG with those only based on CNN descriptors, but also tested our final `combined' PoT representing using both appearance descriptors and motion descriptors (i.e., HOF and MBH) by replacing CNN with HOG.}

The idea was to compare two appearance descriptors (CNN vs. HOG) in representing first-person videos, and confirm that our PoT is appropriately taking advantage of CNN descriptors which is supposed to perform superior to HOG. Figure \ref{fig:cnn-vs-hog} illustrates the results obtained with the two datasets we use. We are able to observe that, with our PoT, CNN descriptors are significantly outperforming HOG descriptors under identical settings.



\section{Conclusion}

We introduced the new feature representation designed for first-person videos: \emph{pooled time series} (PoT). Our PoT was designed to capture entire scene dynamics as well as local motion in first-person videos by representing long-term/short-term changes in high-dimensional feature descriptors, and it was combined with four different types of per-frame descriptors including CNN features. We evaluated our PoT using two public first-person video datasets, and confirmed that our PoT clearly outperforms previous feature representations (i.e., BoW and IFV) as well as the other state-of-the-art video features.


{\flushleft\textbf{Acknowledgement:} The research described in this paper was carried out at the Jet Propulsion Laboratory, California Institute of Technology, under a contract with the National Aeronautics and Space Administration. This research was sponsored by the Army Research Laboratory and was accomplished under Cooperative Agreement Number W911NF-10-2-0016.}

\bibliographystyle{ieee}
\bibliography{egocentric}

\end{document}